# A Novel Approach for Machine Learning-based Load Balancing in High-speed Train System using Nested Cross Validation

İbrahim Yazıcı[1], and Emre Gures[2]

[1]Türk Telekom R&D Department, Acıbadem, Istanbul, Turkey
[2]Faculty of Electrical Engineering Czech Technical University in Prague, Prague, Czech Republic
Email: [1]ibrahim.yazici@turktelekom.com.tr, [2]guresemr@cvut.cz

*Abstract*— Fifth-generation (5G) mobile communication networks have recently emerged in various fields, including high-speed trains. However, the dense deployment of 5G millimeter wave (mmWave) base stations (BSs) and the high speed of moving trains lead to frequent handovers (HOs), which can adversely affect the Quality-of-Service (QoS) of mobile users. As a result, HO optimization and resource allocation are essential considerations for managing mobility in high-speed train systems. In this paper, we model system performance of a high-speed train system with a novel machine learning (ML) approach that is nested cross validation scheme that prevents information leakage from model evaluation into the model parameter tuning, thereby avoiding overfitting and resulting in better generalization error. To this end, we employ ML methods for the high-speed train system scenario. Handover Margin (HOM) and Time-to-Trigger (TTT) values are used as features, and several KPIs are used as outputs, and several ML methods including Gradient Boosting Regression (GBR), Adaptive Boosting (AdaBoost), CatBoost Regression (CBR), Artificial Neural Network (ANN), Kernel Ridge Regression (KRR), Support Vector Regression (SVR), and k-Nearest Neighbor Regression (KNNR) are employed for the problem. Finally, performance comparisons of the cross validation schemes with the methods are made in terms of mean absolute error (MAE) and mean square error (MSE) metrics are made. As per obtained results, boosting methods, ABR, CBR, GBR, with nested cross validation scheme superiorly outperforms conventional cross validation scheme results with the same methods. On the other hand, SVR, KNRR, KRR, ANN with the nested scheme produce promising results for prediction of some KPIs with respect to their conventional scheme employment.

## I. INTRODUCTION

The popularity of high-speed railways is on the rise, driven by their many benefits, including lower energy consumption, increased transport capacity, reduced pollution, time efficiency, safety, and comfort [1], [2]. To meet growing demand for mobile wireless services, broadband communication has become essential in high-speed train systems [3], [4]. The utilization of millimeter wave (mmWave) frequency bands in fifth-generation (5G) and beyond (B5G) cellular networks offer the potential for accommodating exponentially increasing data rates through new spectrum bands [5]. Nevertheless, mmWave bands come with challenges such as high path loss and susceptibility to rapidly changing channel conditions. To this end, dense deployment of mmWave cells is needed to ensure sufficient coverage and capacity, and the integration of edge computing further complicates these challenges.

High-speed rail networks experience frequent handovers (HOs) due to the dense deployment of 5G mmWave base stations (BSs) and the high speeds of trains. This results in increased signalling overhead and latency, highlighting the significance of effective HO management and resource allocation during user association. Load balancing through user association has emerged as a promising approach to handle higher data rates, manage cell congestion, and optimize wireless resource allocation across multiple links [6], [7]. Load imbalance leads to an unfair distribution of data rates among users, resulting in inconsistent quality of experience (QoE) for users. To handle this issue, load balancing is employed to transfer the users from overloaded cells to lightly-loaded cells, ensuring a fair distribution of the network load among cells.

HO is a critical process in high-speed rail networks, providing seamless call switching between cells as the users move within the cellular coverage area. Proper HO management is vital to maintain quality of service (QoS), prevent reductions in service interruptions and average throughput. Key HO control parameters (HCP), such as HO margin (HOM) and time-to-trigger (TTT), play a crucial role in load balancing. HOM adjusts the cell coverage by modifying pilot power value of the cells, making the underloaded cells more appealing for user association. While this improves overall system traffic handling and increases user throughput, it can lead to lower signal-to-interference-plus-noise ratio (SINR) for users at the cell edge. As a result, setting the HOM value appropriately becomes crucial for ensuring efficient HO processes. Another important parameter is the TTT interval, which regulates duration for testing received signal strength before executing a HO. Reducing the TTT time allows for earlier handovers from the overloaded cells to underloaded cells, leading to improved average throughput and a lower call dropping ratio (CDR). However, low TTT settings may increase HO probability (HOP) and HO ping-pong (HOPP), while high TTT settings can impede offloading from overloaded cells, thereby causing delays in HOs for users experiencing poor communication quality.

This study presents insight into deployment of ML methods for KPIs' prediction for the high-speed train scnenario in 5G and B5G mobile networks. ML perdiction with nested cross validation may enhance the prediction performance by increasing generalization ability of ML methods. This will pave ways for future reserach and applications. Since the performances of the boosting methods have greatly improved with nested validation scheme, more efficient load balancing algorithms may be proposed with this ML modelling of the high speed train system scenarios. In addition, this ML modeling may be applied for different KPI prediction scenarios in 5G and B5G network systems as well.

This paper is structured as follows: Section II surveys the relevant literature on this topic. Section III introduces the system and simulation models. Section IV provides an overview of the ML techniques employed. Section V discusses data collection, processing, and the practical application of our methods to address the problem. Section VI presents the outcomes obtained by employing the ML methods. Finally, Section VII concludes the paper with a summary of the key findings and concluding remarks.

## II. RELATED WORKS

Numerous mobility management algorithms utilizing machine learning (ML) in train networks are also available in the literature. Ref. [8] presents a HO decision algorithm based on Bayesian regression for Long Term Evolution for Railway (LTE-R) networks. The algorithm predicts crossing time of cell boundaries and determines initiation of HO process. However, the algorithm makes idealized assumptions and neglects the impact of multiple factors on HO outcomes. Ref. [9] proposes a parameter-adaptive HO mechanism for LTE-R systems that can be adapted to 5G. The mechanism first constructs a discrete TD value cube, which represents the handover performance and network performance for different combinations of speeds and HO parameters. The function approximation method is then used to approximate the continuous HO parameter selection problem, resulting in a continuous TD value cube and the corresponding performance cubes. The experimental results show that the proposed mechanism can also find the optimal handover parameters to improve HO performance and network performance. In Ref. [10], an improved HO decision strategy is proposed based on Q-learning algorithm to reduce the number of unnecessary HO and improve the network performance. The proposed scheme considers four parameters including HO cost, signal-to-interference-plus-noise ratio (SINR), received signal reference power (RSRP) and time of stay (ToS) as rewards, and uses Q-learning algorithm to link the users' current decision to the long-term benefit to improve the HO decision. The simulation results show that the proposed scheme outperforms the conventional approach.

Existing HO algorithms have some limitations in 5G networks. These limitations are due to the fact that some of them are derived from LTE technology. Since LTE and 5G have different features and requirements, these algorithms are not optimally efficient for 5G networks. To address this, further research is required to explore various mobility and deployment scenarios specific to 5G and develop algorithms tailored to its characteristics.

The aforementioned algorithms have been predominantly designed based on a centralized optimization model, which may result in increased handover issues for certain users [11]. However, not all mobile users require simultaneous or identical optimization, which becomes more crucial considering the limited coverage of 5G BSs and the necessity to support high mobility speeds of up to 400 km/h. Additionally, the optimization scope of existing algorithms is another challenge. Some algorithms only focus on optimizing a single HCP, such as HOM or TTT, as observed in [8], [9], [12]. Consequently, there is a need to consider the optimization of all relevant HCPs.

## III. NETWORK AND SIMULATION MODELS

In Ref. [13], a simulation environment, developed in MATLAB 2020b, models a B5G network with microcells and railway network scenario. The network layout consists of thirty-nine evolved NodeBs (eNBs) with three sectors for each cell, following LTE-Advanced Pro 3rd Generation Partnership Project (3GPP) Rel. 16 specifications. The simulation dynamically adjusts the number of hexagonal cells based on the simulation time. The simulation incorporates a frequency reuse factor of one.

The simulation model mimics the random generation of traffic load, accounting for admission control functionality and omnidirectional antenna communication. To simulate high-speed train movement, Directional Mobility Model (DMM) is employed for all measured users, restricting their movement to one direction on parallel paths at a speed of 400 km/h, and the simulation updates user movements at a periodic interval of 40 ms. For performance evaluation, 15 users are selected, and their average values represent the measurements throughout the study. In the initial simulation cycle, users are randomly generated to achieve independent measurements, enhancing result accuracy. Average values for various metrics, including Load Level (L), Throughput (T), CDR, Spectral Efficiency (SE), HOP, HOPP, and RLF, are calculated in each simulation cycle.

*1) Data generation*

The dataset utilized in this study was generated using MATLAB 2020B software, specifically tailored for the B5G network and incorporating microcell and railway network scenario based on the simulation settings used in Ref [13]. The simulation generated 528 samples by systematically varying the fixed HOM and TTT values, in accordance with the specifications outlined by 3GPP. The HOM values ranged from 0 to 16 dB with increments of 0.5 dB, while the fixed TTT intervals followed the predefined values: 0, 40, 64, 80, 100, 128, 160, 256, 320, 480, 512, 640, 1024, 1280, 2560, and 5120 ms.

This dataset encompasses a wide range of measurements to provide a comprehensive analysis, considering additional KPIs and evaluating network performance from diverse perspectives. In addition to load balancing-related KPIs such as L, CDR, T, and SE, the analysis includes performance metrics related to mobility robustness optimization (HOP, HOPP, and RLF) to address the challenges posed by high-speed train scenarios.

*2) Handover Decision*

The main objective of load balancing is to achieve a balanced distribution of network traffic across cells by redirecting user traffic from overloaded cells to underloaded cells. This is achieved by introducing a bias value to pilot power of a target cell. Load balancing function increases the HOM, prompting early HOs to other cells when a serving cell is overloaded. By prioritizing the underloaded target cells with this biased power during a HO decision stage, network system efficiently serves more traffic, leading to higher throughput for users. However, this unanticipated increase in HOM may result in RLFs and HOPP, adversely affecting user QoS. To overcome these challenges, an automated model is needed to predict HOM settings independently for each user based on cell load. Furthermore, conventional A3 HO event, which relies on a power-based margin, is triggered when the target cell's power surpasses that of the serving cell for one TTT period. The event can be expressed as follows:

$$RSRP_t + HOM > RSRP_s \qquad (1)$$

Variables $RSRP_s$ and $RSRP_t$ denote RSRP values of the serving cell and the target cell, respectively. In the case where the load balancing function is not activated, indicating that the serving cell is not overloaded (i.e., the load level is below 65%), the HOM value is assigned to be zero [14], [15]. To achieve load balancing and mitigate the occurrence of HOPP and HOP, an adjustment in the target cell selection process is implemented. The target cell is determined based on a restricted list of neighbouring cells, which is defined according to the load level criterion. Within this list, the target cell is chosen as the one with the highest RSRP. This adjustment ensures that users are handed over to underloaded target cells, provided that the necessary conditions are met.

IV. SYSTEM PERFORMANCE PREDICTION THROUGH MACHINE LEARNING

Recently, ML method deployments across different fields have gained attention, and this is the case in communications field as well. Ease of the deployments and promising result yielding of ML methods make ML deployment attractive for many studies. In this paper, we aim to predict the performance of a high speed train system scenario through some ML methods. In the deployments, outputs of the high speed train system dependent on HOM and TTT input values are predicted. This prediction by the ML deployments may be useful for both load prediction and Handover Control Parameters (HCPs) optimization as the further study [13].

ML consists of three types of learning according to their learning mechanisms that are supervised, unsupervised, and reinforcement learning types. In our deployments, supervised learning will be utilized in the context of ML types. On the other hand, supervised learning consists of two tasks which are classification and regression. Classification task is performed by classifying the outputs according to their categories with respect to some input values whereas regression task is performed by predicting a real valued output with respect to some input values. In this paper, regression task is used to predict multiple outputs with respect to HOM and TTT input values, and the outputs are L, T, CDR, SE, RLF, HOP, and HOPP. To achieve our aim as predicting the outputs, we utilize some efficient and prominent ML methods which are SVR, ABR, GBR, CBR, ANN, KRR, and KNNR. These methods were used in a previous study [13] with the same data utilized in this paper as well, and this study aims to extend the previous work by approaching to the same problem with a different scheme, so that we use the same ML methods in the previous study for better comparison of the results. In the next subsections, brief information about the utilized ML methods are given.

A. *Support Vector Regression*

Support Vector Machine (SVM) is one of the prominent ML method based on statistical learning theory [16]. The motivation behind the method is to maximize the gap between two different categorical outputs by separating them in a linear or non-linear manner. In the non-linear manner, feature space are mapped into a higher dimensional space with kernel trick so as to make classification that is not possible with linear mapping of the feature space. The core concept of SVM method is that this kind of the mapping of feature spaces. By this way, SVM problem is framed as a constrained optimization problem, then it is solved by quadratic optimization. Beside performing classification task, SVM performs regression tasks as well. It is called Support Vector Regression, and works with real valued outputs rather than categorical outputs.

B. *AdaBoost Regression*

AdaBoost method is a type of ensemble learning that is used for both classification and regression tasks [17]. In the context of ensemble learning, there are two kind of ensemble learning which they are called bagging and boosting. In the learning process of bagging strategy, model construction is based on homogeneous weak learners' learning process which is independently parallel from each other, and the learned models are combined to determine average of the model. On the other hand, in the learning process of boosting strategy, model construction is based on homogeneous weak learners' learning process which is sequential and adaptive contrary to bagging strategy. AdaBoost method is a kind of boosting method, and the method performs both classification and regression task efficiently.

C. *Gradient Boosting Regression*

Gradient Boosting Regression is a type of boosting algorithm, and it is a modified version of adaptive boosting (AdaBoost) method. In the learning process, strong learners are formed from weak learners by compensating the weakness of weak learners which the weak learners evolve into strong learners in a parallel manner [13], [18], [19]. Boosting is framed as an optimization problem which aims to minimize loss function through gradient type optimization procedure [13].

D. *CatBoost Regression*

Cat Boosting method is a novel type of boosting method proposed by Yandex research. It handles categorical features effectively along with continuous-valued features in a dataset, and provides scalable and GPU computations with the use of large datasets [20].

## E. Artificial Neural Network

Artificial Neural Network (ANN) is a kind of ML method that mimics learning process of human brains. In this sense, it uses some layers formed by neuron-type nodes, and performs learning process among the interactions of these neurons. With recent developments, there are many types of neural networks which oftentimes achieve superior performance with respect to human-level performance. Recurrent Neural Networks, Convolutional Neural Networks, Transformers, Long-Short Term Memory, Gated Recurrent Units are landmark examples of novel neural network types. Simply, a feed-forward ANN, used in this paper as well, consists of one input, one hidden, and one output layer. Deep feed-forward ANNs use more than one hidden layer in their architectures which affects overfitting or underfitting of the ANN model. In the hidden layer(s) and in the output layer, learning is facilitated by activation functions. The ANN performs both regression and classification tasks, and it is determined by the type of the utilized activation function in the output layer.

## F. Kernel Ridge Regression

Kernel Ridge method performs nearly the same operations with SVM by kernel trick which transforms feature space of a dataset into a higher dimensional space to better represent the dataset. A model is constructed through kernel trick to a parametric model, this is the ridge regression [13], [21]. Learned model by kernel ridge method is non-sparse meaning that the solution vector is solely dependent on all training inputs. Even kernel ridge and SVM methods are similar to each other, they differ in a manner that they utilize different loss functions. While the former method utilizes squared error loss, the latter utilizes epsilon-insensitive loss [13].

## G. K-Nearest Neighbor Regression

KNNR is a kind of non-parametric ML method which fits model according to neighbourhood of pre-determined $K$ numbers of neighbours [22]. $K$ neighbours' values are averaged to produce output for regression task of KNNR, and a step function leading to a smoother model fitting is used in model construction phase of KNNR [23].

## V. MACHINE LEARNING DEPLOYMENT

In this section we provide ML deployment stages. The deployments stages consist of data collection, data processing, and model training.

### A. Data Collection

Utilized dataset is taken from a recently published paper [13]. The dataset simulated microcells and railway network scenarios comprises of 528 data points which each data point has 9 features. 2 of the features (HOM and TTT) are inputs, and the rest of them (L, T, CDR, SE, HO PP, HOP) are used as outputs.

### B. Data Processing

In ML deployments, some models are sensitive to feature scaling that affects obtained results, and convergence speed. Tree-based methods do not need to be normalized as they are not sensitive to the variance in a dataset. However, ANN needs for a scaled dataset, and it had better use scaled inputs for SVR as well. For deployed ML methods excluding tree-based methods, we use zero-mean normalization for features given as in Equation 1:

$$x'_i = \frac{x_i - \mu}{\sigma} \quad (1)$$

where $x'_i$ and $x_i$ denote normalized and original values of $i^{th}$ data point, respectiveky. On the other hand, $\mu$ and $\sigma$ represent mean and standard deviation of relevant feature, respectively.

### C. Model Training

In the deployment in this paper, we use nested cross validation rather than conventional cross validation for training of ML methods. In the conventional scheme, ML method utilized tunes its parameters and evaluates model performance with the same data by using the same validation dataset, thereby leaking the learned information from the same data into the model. Chosen parameters which maximize the performance of non-nested cross validation makes the model biased to utilized dataset, and the model produces overly-optimistic score. As a result, the model overfits the dataset. However, this is not problem for nested cross validation scheme since it uses different subsets from the dataset for evaluating the model and choosing the parameters for model. Nested cross validation makes use of several parts of train/test/validation splits in its operations. In the inner loop of this scheme, each training set is utilized for fitting a model to maximize the score, and then the score is directly maximized via using validation set. In the outer loop, test dataset splits are used for estimating generalization errors. In the inner loop, the information leakage of model fitting into the parameter selection of this model is prevented as contrary to non-nested cross validation scheme. Hence the results are less affected by overfitting issue. Through using the latter scheme of the cross validation, this paper extends the previous paper [13].

In Ref. [13], train-test split ratio was 85%-15%, and %10 of the train set was allocated to validation split. Moreover, 10-fold cross validation was applied for the deployment of ML methods in the paper. In our deployment splits are by 6-fold and 4-fold for outer loop and inner loop, respectively,

We use multi-input multi-output (MIMO) prediction strategy for output (KPIs) predictions of the train system. For performance comparisons of the ML methods used, we utilize Mean Absolute Error (MAE) and Mean Square Error (MSE) given in Equations 2-3:

$$MAE = \frac{1}{N}\sum_{i=1}^{N}|y_i - \hat{y}_i| \quad (2)$$

$$MSE = \frac{1}{N}\sum_{i=1}^{N}(y_i - \hat{y}_i)^2 \quad (3)$$

In Equations 2-3, $N$ denotes the size of dataset used, $y_i$ and $\hat{y}_i$ correspond to actual and predicted values of $i^{th}$ example, respectively.

## VI. RESULTS AND DISCUSSIONS

We present obtained results of ML deployment, and discuss the results with previously obtained results with the same data in Ref. [13] which used conventional cross validation. The obtained results are compared in terms of MAE and MSE metrics. Performance of ML methods with nested cross validation scheme are discussed. In addition, discussion of the performance of ML methods with nested cross validation denoted by * with respect to conventional cross validation results. The results are presented in Tables 1-2.

TABLE 1.  MAE RESULTS

| | MAE (%) | | | | | | |
|---|---|---|---|---|---|---|---|
| | L | T | CDR | RLF | SE | HOPP | HOP |
| ABR | 0.48 | 1.57 | 36.92 | 8.9 | 36.73 | 0.48 | 1.5 |
| ABR* | **0.28** | **0.35** | **7.5** | **1.73** | **7.5** | **0.048** | **0.23** |
| GBR | 0.3 | 1.45 | 35.40 | 8.2 | 35.40 | 0.48 | 1.48 |
| GBR* | **0.02** | **0.006** | **0.09** | **0.03** | **0.09** | **0.001** | **0.03** |
| CBR | 0.3 | 1.44 | 35.68 | 8.2 | 35.68 | 0.48 | 1.48 |
| CBR* | **0.04** | **0.054** | **2.22** | **0.32** | **2.22** | **0.012** | **0.4** |
| SVR | 1.79 | **5.92** | 207.2 | **33.3** | 207.2 | **0.85** | **3.8** |
| SVR* | **1.76** | 6.14 | **204.3** | 34.63 | **204.3** | 1.38 | 5.02 |
| MLP | 1.13 | **1.65** | 158.8 | **9.68** | 157.9 | **0.45** | **1.08** |
| MLP* | **1.01** | 1.8 | **151.2** | 10.2 | **149.8** | 0.56 | 1.22 |
| KNNR | 1.09 | 2.2 | 113.14 | 12.36 | 113.14 | **0.5** | **1.51** |
| KNNR* | **0.72** | **1.88** | **62.45** | **10.65** | **62.45** | 0.66 | 1.9 |
| KRR | 0.91 | **1.63** | 114 | **9.22** | 114 | **1.08** | **2.62** |
| KRR* | **0.83** | 1.81 | **106.9** | 10.31 | **106.87** | 1.54 | 3.7 |

In the deployment stages, we first get the dataset, and make some feature scaling for some of ML methods used as mentioned in previous sections. This is followed by model construction for each method. To this end, we use nested cross alidation scheme to get more stable and robust generalization errors. In our deployment for nested cross validation in this paper, we use 6-fold split for outer cross validation set, and 4-fold split for inner cross validation set. In non-nested cross validation scheme, 10-fold cross validation was made by the ML methods used in Ref. [1-AEJ]. Then, we deploy ML methods and compare obtained results with respect to non-nested cross validation deployment version of the same methods in terms of MAE and MSE metrics.

TABLE 2.  MSE RESULTS

| | MSE (%) | | | | | | |
|---|---|---|---|---|---|---|---|
| | L | T | CDR | RLF | SE | HOPP | HOP |
| ABR | 0.003 | 0.036 | 26.53 | 1.15 | 26.41 | 0.006 | 0.051 |
| ABR* | **0.002** | **0.003** | **1.3** | **0.07** | **1.3** | **4e-5** | **9e-4** |
| GBR | 0.002 | 0.033 | 25.25 | 1.07 | 25.25 | 0.006 | 0.049 |
| GBR* | **8e-5** | **2e-5** | **0.006** | **0.0005** | **0.006** | **4e-7** | **1e-6** |
| CBR | 0.002 | 0.033 | 25.32 | 1.08 | 25.32 | 0.006 | 0.049 |
| CBR* | **5e-5** | **0.0001** | **0.15** | **0.005** | **0.15** | **13e-5** | **13e-5** |
| SVR | 0.089 | **0.492** | **1296** | **15.67** | **1296** | **0.05** | **0.45** |
| SVR* | **0.088** | 0.54 | 1329 | 17.12 | 1329 | 0.086 | 0.7 |
| MLP | 0.019 | **0.053** | 380.3 | **1.85** | 377.25 | **0.005** | **0.036** |
| MLP* | **0.017** | 0.060 | **340.93** | 1.9 | **334.37** | 0.012 | 0.054 |
| KNNR | 0.039 | 0.079 | 409.49 | 2.5 | 409.49 | **0.006** | **0.055** |
| KNNR* | **0.026** | **0.086** | **256.83** | **2.8** | **256.83** | 0.04 | 0.21 |
| KRR | 0.013 | **0.051** | 235.72 | **1.67** | 235.72 | **0.029** | **0.173** |
| KRR* | **0.012** | 0.055 | **201.6** | 1.8 | **201.6** | 0.045 | 0.26 |

The results are presented in Tables 1-2, and the best performance of each method and each feature is written in bold in the tables. Performance differences are also given in tables which the top-2 best performances are compared. Boosting methods that use nested cross validation yield promising results when compared to their non-nested cross validation counterparts. However, SVR, MLP, KNNR, and KRR methods with nested scheme do not produce fully superior results over the results of non-nested scheme.

In method-level analysis in Table 1, ABR has improved the non-nested scheme results by 0.42, 0.78, 0.8, 0.81, 0.8,0.9, 0.85 for load level, throughput, call drop ratio, radio link failure, spectral efficiency, handover ping-pong probability, and handover probability, respectively. GBR method has also produced promising results with respect to its non-nested scheme counterpart in improving the results. It has improved the results for the same feature values mentioned in order by 0.93, 0.99, 0.99, 0.99, 0.99,0.99, 0.98. For another boosting method, CatBoost, the results have been improved by 0.87, 0.96, 0.94, 0.96, 0.94, 0.975, 0.73 for the features in order. For SVR method, the nested scheme produce better results for load level, call drop ratio, and spectral efficiency by 0.11, 014 and 0.14 in order. However, it stays behind the non-nested scheme for the rest of the outputs. MLP and KRR show the same characteritics with SVR which MLP has improved the results with the nested scheme by 0.11, 0.048, and 0.051 for the same features with SVR. KRR has improved the results by 0.08, 0.063, and 0.062 for the same features with SVR. However, there is no improvement for the rest of the outputs with MLP and KRR. KNN method has differed from the other methods in prediction. It has improved the result for LL, T, CDR, RLF, SE in order by 0.34, 0.15, 0.45, 0.14, 0.45. Overall, it is apparent that the nested scheme improves the results with respect to its non-nested counterpart as per the obtained MAE results. The boosting methods are the most beneficial ones from these improvements.

As per MSE results given in Table 2, all the boosting methods has again yielded better results than their non-nested counterparts. CBR and GBR nearly have had performance gain with respect to its non-nested scheme, and the best performance gain having method (GBR nested denoted as GBR*) are given in Figures 1-2. The performance improvements for ABR in LL, T, CDR, RLF, SE, HOPP, and HOP in order are 0.33, 0.92, 0.95, 0.94, 0.95,0.99, 0.98. The improvements for GBR for the same features are 0.96 , 0.99, 0.99, 0.99, 0.99, 0.99, 0.99. The improvements for CBR for the same features are 0.98, 0.99, 0.99, 0.99, 0.99, 0.98, 0.99. On the other hand, the nested scheme of the other methods used does not provide promising results with respect to their non-nested counterparts. The nested scheme with SVR only passes in LL with 0.01 improvement.

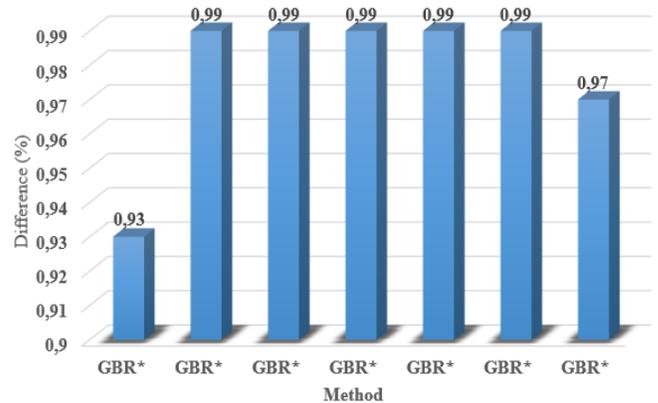

Fig. 1.  Best performing method for each output in MAE metrics.

With MLP, L, CDR and SE results have been improved by 0.11, 0.12, and 0.11 in order. The same results with MLP are valid for both KNNR and KRR. With KNNR, only LL, CDR and SE results among all outputs have been improved by 0.33, 0.35, and 0.35 in order. With KRR, LL, CDR and SE results have been improved by 0.077, 0.145, and 0.145 in order. Overall, the nested scheme has apparently improved the results for the boosting methods.

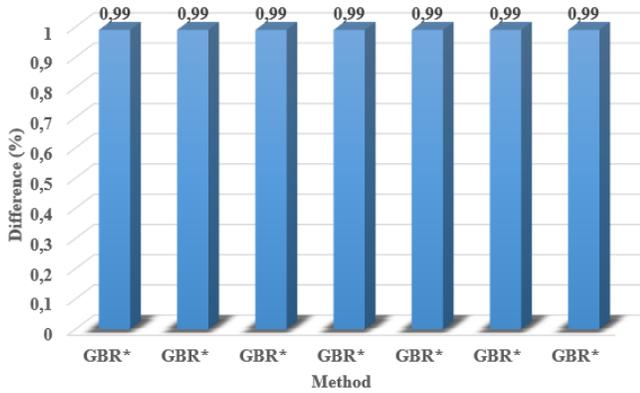

Fig. 2. Best performing method for each output in MSE metrics.

## VII. CONCLUSIONS

In this paper, a new approch to model mobility management of the high-speed train system behaviour based on ML from the perspective of predictive analytics is utilized. With ML prediction, we aim to provide system modelling for the high-speed train scenario in a 5G network to predict several outputs for different HOM and TTT values. The study may also be used for different mobility scenarios with deployed settings in this paper, and may produce promising results for different scenarios in 5G mobile communication systems. In the deployment stage, dataset used is processed for the relevant ML methods, and this is followed by deployment of ML methods with nested cross validation scheme. GBR, ABR, CBR, SVR, MLP, KNNR, and KRR are the methods utilized in the paper. Results of the methods with nested cross validation scheme are compared with its non-nested scheme with the same methods in a previous study in terms of MAE and MSE metrics [13]. According to the results, the boosting methods achieve superior performance with respect to its non-nested scheme in predicting all outputs. On the other hand, SVR, MLP, KNNR, and KRR methods achieve the best results for some outputs with nested cross validation scheme. Hence, nested cross validation scheme has produced superior performance for the boosting methods with nested scheme, however, it has not produced fully superior performance with SVR, KNNR, KRR, and MLP.